# Aim in Climate Change and City Pollution

Pablo Torres, Beril Sirmacek, Sergio Hoyas, and Ricardo Vinuesa

## Contents

P. Torres · S. Hoyas
Instituto Universitario de Matemática Pura y Aplicada,
Universitat Politécnica de Valencia, Valencia, Spain
e-mail: pablotg@kth.se

B. Sirmacek
Smart Cities, School of Creative Technologies, Saxion University of Applied Sciences, Enschede, The Netherlands
e-mail: b.sirmacek@saxion.nl

R. Vinuesa (✉)
FLOW, Engineering Mechanics, KTH Royal Institute of Technology, Stockholm, Sweden
e-mail: rvinuesa@mech.kth.se

**Abstract**

The sustainability of urban environments is an increasingly relevant problem. Air pollution plays a key role in the degradation of the environment as well as the health of the citizens exposed to it. In this chapter we provide a review of the methods available to model air pollution, focusing on the application of machine-learning methods. In fact, machine-learning methods have proved to importantly increase the accuracy of traditional air-pollution approaches while limiting the development cost of the models. Machine-learning tools have opened new approaches to study air pollution, such as flow-dynamics modeling or remote-sensing methodologies.

## 1 Introduction

Urban areas are at the center of the current climate-change debate. Starting at the end of the industrial revolution, urban areas have been growing at an accelerated rate. By 2050 the European Commission expects 70% of the global population to live in urban environments [4]. Thus, it is clear that cities play and will continue to play a major role in our society. Cities are responsible for a great share in the acceleration of climate change. Mainly due to anthropogenic activity, cities act both as concentrators and diffusers of contaminants at the expense of not only the city itself but also of neighboring areas. In fact, ambient air pollution is responsible for 790,000 deaths in Europe [11], as shown in more detail in Fig. 1. Furthermore, the vast majority of cities tend to act as the so-called urban-heat-islands (UHI), modifying the natural thermal dynamics of neighboring areas. For those reasons among others, cities are expected to play a key role in the policies and endeavors aiming to reverse the effects of climate change within the next decade [18].

To deal with pollutant dispersion within the urban environment, one inevitably needs to consider the dynamics of urban flows. Environmental sciences tend to rely on the use of models to study and predict the behavior of atmospheric flows. Models can provide very useful information when studying atmospheric phenomena at a very general level or when dealing with specific applications. However, to understand the flow within an urban environment as well as the inner relation to pollution mechanisms, modeling appears to be limited. In fact, urban flows are characterized by their complexity, thus severely difficulting the use of models to study the effects on pollutant dispersion or thermal problems. The experimental approach, although it can bring very useful insight into specific applications, typically does not provide information on the physical mechanisms driving urban flows [17]. Numerical simulations are also typically used in the study of turbulent flows and thus can be applied within the context of urban environments. Numerical approaches vary in terms of accuracy and complexity. The so-called Reynolds-averaged Navier-Stokes simulation (RANS) is widely used for its simplicity and reduced computational cost. Unfortunately, RANS methods exhibit important limitations when dealing with complex turbulent flows such as urban flows, thus limiting its use within the context of urban environments. In this way, urban flow simulations have to rely on more accurate approaches, such as direct numerical simulations (DNS) and large-eddy simulations, which have high computational cost. In addition, the domains considered in urban environments are typically large and complex, which again increase the computational cost.

The use of artificial intelligence (AI), and in particular of machine-learning methods, can improve the aforementioned limitations of the computational approach in the study of urban flows. In fact, using machine learning techniques one can produce turbulent models that are significantly more precise than the models historically used in computational fluid dynamics (CFD) without reaching the computational cost of DNS and LES methods. Those models would increase the forecasting capabilities of numerical methods in the study of urban flows. The next section will focus on explaining the state of the art of the aforementioned methods as well as the lines of study currently being explored.

## 2 Machine-Learning Methods in the Study of Urban Pollution

Machine-learning (ML) techniques have been widely used in many fields of physics and scientific research. In fact, ML has proved to be very effective in extracting underlying patterns and correlations within complex physical processes. The present section will focus on introducing the different ML approaches in the study of urban

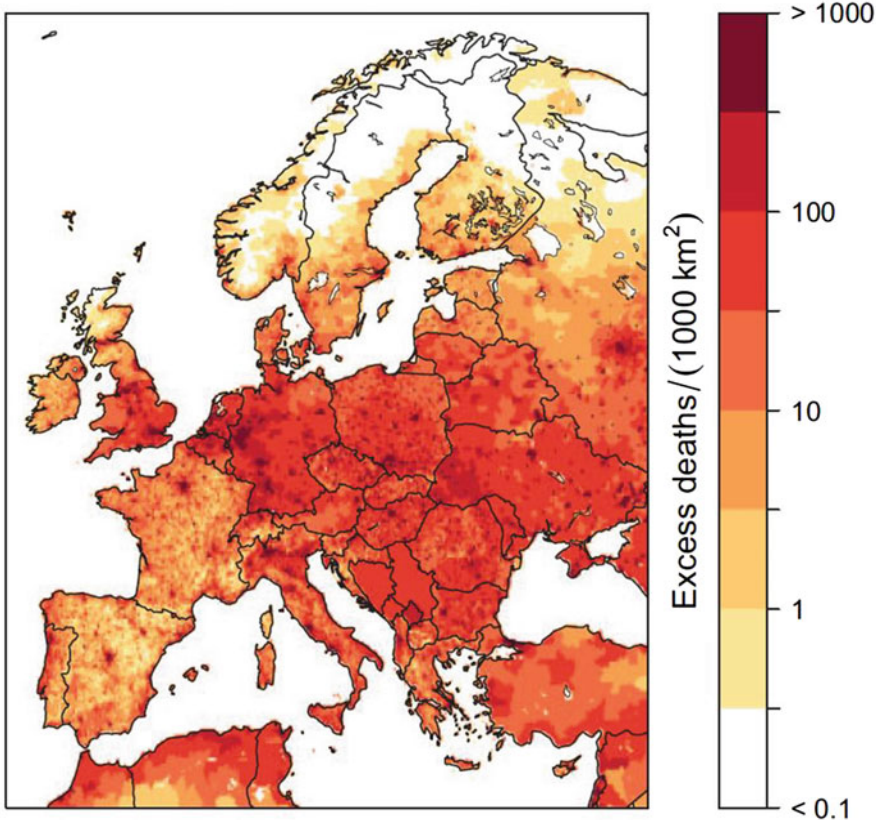

**Fig. 1** Distribution of annual excess mortality from cardiovascular diseases due to air pollution in Europe. (Figure reproduced from Lelieveld et al., with permission from the publisher (Oxford University Press))

pollution. When dealing with urban environment pollution, two approaches are found. On the one hand, some studies focus on modeling pollutant dispersion as an independent physical phenomenon. Those models have been widely used and typically rely on empirical relations. Using ML methods, the aforementioned models can be significantly improved, as we will see later on. On the other hand, many studies rely on flow dynamics to characterize pollutant dispersion. This approach typically consists of solving the behavior of the flow in an urban environment in order to model the behavior of the pollutant. Several numerical methods have been widely used to solve urban flow dynamics (e.g., RANS and LES) but they are either too simplistic or too expensive in terms of computational cost. In this particular area, ML methods are being developed in order to reproduce flow dynamics without having to incur such computational costs.

## 2.1 ML Methods in Air-Pollutant Modeling

Air-pollution models have been widely used in order to analyze and control air quality in cities. Air quality is typically measured in terms of the particulate matter (PM) concentration, which basically measures the amount of particles found in the air. PMs are classified in terms of the size of the particulate. For instance, PM 10 refers to particles with diameters of 10 micrometers or smaller. In the present section we will focus on PM 2.5 as it is common to urban environments and it has been associated both with adverse health effects [21] and climate forcing [1]. Pollutant models have traditionally relied on linear regression techniques to capture the underlying mechanisms of pollution. This approach is known to be limited, since urban-flow phenomena are characterized by their complexity, that is, the presence of nonlinearities and highly correlated effects. Complex phenomena are known to be challenging to linear methods as they tend to misrepresent important underlying relations [20]. ML arises as a solution to this problem and thus enhances air-pollution models. The fundamental idea behind ML pollutant models lies on using data, typically obtained with on-site measurements, to train the model such that it can predict a certain quantity. In the case of pollution models, the output of the model is normally pollutant concentration, that is, PM concentration. In the next lines we will present the most common ML algorithms applied to urban pollution models and compare them in terms of performance and accessibility.

The random-forest (RF) algorithm is a popular ML method to solve classification and regression problems. It is based on a classical decision-tree algorithm widely used in statistics and ML, but it builds each individual tree from random subsamples of the original data. In this way, each subsample of the original data produces a tree and thus produces a predicted response (or a predicted class if dealing with a classification problem). The final output of the model is determined by the average of each of the predictive responses (or classes) determined by each of the data subsamples. The use of a diverse database increases the probability of success [6], thus being very convenient to analyze complex phenomena which typically have important disparities within batches of data. In addition, thanks to the averaging process the algorithm can handle missing values in the dataset [3]. The accuracy of the method is dependent on the strength of each individual tree and the dependency of each subsample with one another [6].

Boosted regression algorithms are typically the result of modifying decision trees regression algorithms in order to enhance performance. For instance, the boosted regression tree (BRT) combines the well-known classification and regression trees (CART) with a large number of single models in order to improve the predictive performance of the model [6]. In fact, the averaging process found in the RF model is replaced with a forward stagewise procedure in which existing trees are left untouched with new tree development using the residual information generated in the previous step of the process [6]. Alternatively, one can find gradient boosting regression methods such as XG Boost, which use gradient boosted decision trees in classification and regression problems, thus enhancing the speed and performance of the algorithm. In addition, XG Boost can be run using parallel trees as well as a cluster of computers during training time [3], which optimizes the method.

Multilayer perception (MLP) regression is part of the artificial neural network (ANN) family of methods. ANN methods are designed to mimic the behavior of the human brain during the learning process, that is, using interconnected synaptic neurons capable of learning and storing information about their environment [14]. The fundamental component of ANNs are neurons, which are described by a linear combination of weighted input signals and an activation function that limits the amplitude range of the output. ANN is built using at least three layers of neurons: the input, hidden, and output layers. The method processes information sequentially throughout the layers starting at the input layer and finishing at the output layer. The input of the system consists of the training data and a goal to fulfill. In this way, the system is trained on historical data to fulfill the predefined objective, learning during the process the underlying relations that lead to the fulfillment of the goal. Note that several training algorithms can be applied to a given network, for example, the back-propagation algorithm which combine forward and backward passes to train the network [14]. However, taking into account the motivation of the present chapter, we will neglect the training algorithm selection and focus exclusively on the performance of the models once trained. Other, more sophisticated types of ANN include recurrent neural networks (RNNs) and convolutional neural networks (CNNs), which have been used in the context of temporal predictions [13] and non-intrusive sensing [7] of turbulent flows. These strategies have high potential when it comes to developing robust air-pollution frameworks.

The aforementioned techniques are widely used in classification and regression problems and thus their application to air-pollution models – which are regression problems – is currently being studied. The present paragraph aims at discussing the performance of the different ML methods introduced early on. To do so, we will compare each model's performance with one another as well as with linear regression methods used in air-pollution models. To measure performance, four metrics are typically used in the related literature: the mean absolute error (MAE), the root mean square error (RMSE), and the mean square error (MSE). The MAE is defined as the sum of the absolute differences between the predicted values and the actual values scaled with the inverse of the number of observations. Similarly, the MSE is defined as the sum of the squares of the differences between the predicted and actual values scaled with the inverse of the number of observations. The RMSE is just the square root of the MSE. Wang et al. [20] used the aforementioned quantities to compare the performance of two ML methods, the ANN and the XG Boost, with the land use regression model (LUR). The authors developed a least-square LUR by fitting an intercept-only model and then adding explanatory variables (one at a time) based on the ranking of their correlation with the log-transformed of the PM 2.5. Some of the explanatory variables include: land uses, length of neighboring roads, distance to major aerial routes, traffic information, meteorology, etc. The models were developed using air-quality data (PM 2.5) collected over the course of 4 weeks between March and June 2019 in downtown Toronto (Canada). The samples were obtained on a non-rainy weekday between 7:00 AM and noon, covering 19 unique corridors in a 4-by-6 km area [20]. The authors explain that both LUR and ML methods present strengths and flaws. For instance, they found that LUR methods were highly dependent on the selection of the explanatory variables, which are directly dependent on the user's subjective judgment and a priori knowledge. In this way, a change in the selected explanatory variables, ceteris paribus, would have an important effect on the performance metrics (MAE, RME, RMSE, etc.). However, when properly fit, the performance of LUR methods significantly improves – even surpassing ML methods – especially when the size of the dataset decreases [20]. In fact, the size of the dataset is a critical limitation of ML methods since large amounts of data need to be processed within the network's layers such that the model can produce accurate results. Nevertheless, the ANN superiority with respect to linear regression models is guaranteed by the universal approximation theorem of functions, which states that a fully-connected multi-layer feedforward neural network with continuous, bounded and nonconstant activation function can act as a universal approximator for any smooth mapping to any accuracy [20]. The superiority of ANNs over traditional linear methods in the context of turbulent flows has also been thoroughly discussed by Guastoni et al. [7]. The authors conclude that ML methods provide opportunities to understand complex underlying relations in air-pollution dispersion processes as well as nonlinear relations between air quality and exogenous variables. Nevertheless, one cannot directly discard regression methods, since they may exhibit very good performance in the cases where local knowledge is available.

Doreswamy and Yogesh [3] presented a similar study where four ML models (RF, BRT, MLP, and CART) were compared in terms of performance. In addition, they also compared ML methods with classical statistical methods such as the linear regression. Once again, performance is analyzed using the error metrics (MAE, RME, RMSE, etc.) compared both in training and testing results. The dataset used in the process was taking hourly measurements of air pollution (PM $_{2.5}$) at 76 stations distributed over Newport (Taiwan) between 2012 and 2017. The obtained dataset was used to train the different ML algorithms and to compare the predicted air-pollution values with the actual measurements taken in the stations. The gradient-boosting regression – a type of boosting-regression algorithm – showed the best performance compared with both classical statistical methods and ML algorithms. In training, the gradient boosting regressor showed an MSE at one order of magnitude lower than the rest of ML methods and more than two orders of magnitudes lower than the linear regression method. In the rest of the metrics, the obtained values were two to three times lower than the rest of methods. In tests, the difference between the methods was not as large, but the gradient-boosting regressor still obtained errors two times lower than the rest of the methods. In general, all the analyzed ML algorithms exhibited errors two times lower than the ones obtained with the linear regression [3], thus supporting the superiority of ML methods.

In conclusion, we have discussed different ML methods that can be used to develop air-pollution models. For the majority of the algorithms the performance was found to be better than that of classical statistical methods, such as the linear regression. In this way, the application of ML techniques to develop air-pollution models appears to be useful as the accuracy of the models is significantly improved. However, it is important to keep in mind that ML algorithms are limited by the amount of data needed during training, thus not being applicable when large batches of data are not available.

## 2.2 ML Methods to Model Flow Dynamics

The physics of the flow determines the dynamics of pollutant dispersion. In the previous paragraph we approached the study of air quality by modeling pollution, that is, using particulate matter. An alternative approach consists of studying the dynamics of the flow in order to determine the dispersion of pollutants. Numerical simulation such as LES and DNS can be very useful in order to study the dynamics of the flow. However, this kind of numerical technique has a very important computational cost which limits its use. In addition, the setup of the simulation is not a simple task, which again increases the development time. ML methods could significantly improve the current state of numerical simulations in urban flows by improving computational cost and by extension computational time. The main idea is to use the data obtained from a numerical simulation to predict the future behavior of the flow. Note that this approach is relatively new and thus the available literature on the matter is limited. Nevertheless, the aforementioned approach is promising, as shown by Srinivasan et al. [13] in the context of recurrent neural networks. Non-intrusive sensing is another area with very high potential for ML-based methods, as shown by Guastoni et al. [8].

Xiao et al. [22] developed a fast-running non-intrusive reduced-order model (NIROM) to predict the behavior of the flow in urban environments. The authors gathered the data obtained from a high-fidelity numerical simulation (LES) in order to create the starting dataset. They used the data to generate snapshots of the flow fields such that it can be processed later on. A singular value decomposition (SVD) was applied simultaneously to all velocity components such that the natural correlation between the components is captured. The result of the process is a series of proper orthogonal decomposition functions, which are then used in a Gaussian process regression (GPD). The GPD consists of applying a linear combination of Gaussian-shape basis functions in order to obtain surface functions, which are one of the necessary inputs to build the network. Using the aforementioned elements, a neural network was trained in order to predict the behavior of the flow governing equations. The main idea is that one uses a dataset obtained using a high-fidelity simulation – which is computationally very costly – to obtain a model (NIROM) that will be able to predict the behavior of the flow, this time without the need of the dataset. The authors used the NIROM to predict the flow around the London South Bank University such that the prediction could be compared with actual measurements of the flow. They found that the NIROM was capable of making predictions beyond the range of the snapshots thus being able to accurately represent the vast majority of the dynamics represented in the high-fidelity model [22].

The main advantage of the NIROM over other ML-methods lies on the small amount of data required to make it function. Recall the gradient-boosting method presented in the previous paragraph, it required a huge amount of data to produce a reliable output. On the contrary, NIROMs are able to produce an accurate result with a smaller database which can be obtained using a numerical simulation. Comparing the NIROM with high-fidelity numerical simulations, the main advantage of the NIROM lies on computational cost. Once the network is trained, the cost of producing new results is very small compared to the cost of running an additional high-fidelity numerical simulation such as a LES or a DNS. Nevertheless, the NIROM here presented, solves the behavior of the flow and thus needs to be adapted to study pollutant dispersion. This can be done either by including an additional equation in the NIROM (passive scalar) or by running an additional model that uses as input the solution of the governing equation of the flow.

## 3  Remote Sensing for Urban Air Observation

### 3.1  Impact of Remote-Sensing Sensors for Monitoring Urban Airflow

The value of remote-sensing data has been noticed in the early urban airflow and urban air-quality studies. Researchers have found opportunities to measure, visualize, and explain urban airflow and quality using remote-sensing images acquired from satellites. Early studies, however, have used the satellite images mostly as a background map in order to visualize the data which is collected from ground sensors. This is mainly because it took relatively longer time to develop higher resolution satellite sensors and to design new algorithms which can extract information about the urban airflow and air quality.

The newer satellite sensors have provided opportunities to researchers to develop algorithms to predict the airflow speed and direction, land surface temperature (LST), and atmospheric aerosol particles or in other words particulate matter (PM) which indicate the micrometer size of the particles that pollute air.

Accurate and real-time measurements of the airflow from satellite sensors still have limitations due to the available sensor specifications, visit frequencies (number of days needed for the satellite trip in the orbit before it returns back to the same observation point again), spatial resolution, and accuracy of the information. On the other hand, measurement of LST and PM can be done more accurately and frequently with the existing satellite sensors. Therefore, in order to talk about the urban airflow and air qualities frequently LST measurements, PM measurements and the Urban-Heat-Island (UHI) effect are discussed. The UHI effect is the condition that describes higher temperatures in urban areas than surrounding areas of less development. An UHI occurs because of the extensive modification of the land surface. Remarkably higher temperatures are seen mainly in urban areas than in suburban and rural areas, because of the high number of the dense obstacles in urban areas.

Even though the existing remote-sensing satellite sensors cannot be used as high-precision urban airflow measurement tools, they can still provide a number of advantages. The most significant of them could be listed as follows:

- The ground sensors can be used for high-resolution airflow and air-quality measurements. However, the uneven distribution and limited installation possibilities create challenges to collect data which can fully represent the air measurements of the all urban area. On the other hand, using the ground measurements as control points, satellite measurements can be used to estimate the LST and PM for the areas where in situ measurements are not available.
- Satellite images can be used to find the optimal sensor positions in order to install the in situ sensors in urban areas.
- Remote-sensing images can help with monitoring very large areas and make it possible to do large-scale urban airflow research possible.
- Most of the time, it is possible to find digital maps of cities which bring information about the street tunnels, building boundaries, vegetation areas. However, satellite images can provide further information (low/high vegetation type, temperature changes of the lakes and other water reserves, building rooftop types, traffic density, etc.). These details which are acquired from the satellite images give chances to highly enhance the information for doing detailed urban airflow and air quality analysis.

## 3.2 Remote-Sensing Data Resources and Analysis Methods

Satellite sensors can be classified as active and passive sensors. Active sensors (i.e., radar, LIDAR, etc.) first send a wave and generate an image by using the measurements of the waves which are scattered back. Passive sensors (i.e., multispectral, hyperspectral, thermal, etc.), however, generate images by using the light which is naturally received by the sensor. An active sensor, radar, can be used to measure wind vectors over the ocean through radar backscatter, which is also called scatterometry technique. The scatterometer on the SEASAT satellite makes it possible to resolve the wind direction within a 180-degree directional ambiguity, which is then resolved by knowledge of the overall atmospheric pressure pattern. The SEASAT scatterometer was an outstanding success, pointing the way toward future measurements of ocean wind speed and direction. Unfortunately this satellite had a massive power failure 6 months after it started operation. A replacement, stand-alone satellite, QuikSCAT, was launched quickly thereafter in 1999. The scatterometer on QuikSCAT is known as SeaWinds.

Passive sensors cannot measure the wind vectors straightforward like active sensors can measure. However, indirectly they allow researchers to estimate the airflow by measuring LST and PM. Therefore, they are frequently used in urban air flow studies. One of the most frequently used satellite sensors is the Terra satellite which was launched in December 1999. Terra is a highly valuable satellite which has been serving over 20 years for observing our earth. Terra satellite, which is approximately the size of a small school bus, carries five different sensors to observe different qualities of our earth. These sensors are: Advanced Spaceborne Thermal Emission and Reflection Radiometer (ASTER), Clouds and Earth's Radiant Energy System (CERES), Multi-angle Imaging Spectroradiometer (MISR), Measurements of Pollution in the Troposphere (MOPITT), and Moderate Resolution Imaging Spectroradiometer (MODIS). ASTER is the only high spatial resolution instrument on the Terra platform. The Advanced Spaceborne Thermal Emission and Reflection Radiometer obtains high-resolution (15 to 90 square meters per pixel) images of the Earth in 14 different wavelengths of the electromagnetic spectrum, ranging from visible to thermal infrared light. Scientists use ASTER data to create detailed maps of land surface temperature, emissivity, reflectance, and elevation. ASTER provides high-resolution images in 14 different bands of the electromagnetic spectrum, ranging from visible to thermal infrared light. The resolution of images ranges between 15 and 90 meters. ASTER data are used to create detailed maps of surface temperature of land, emissivity, reflectance, and elevation [5]. Because the surface UHIs are typically characterized by (LST), it makes sense to measure LST from remotely sensed data to study UHI effects.

With its sweeping 2330-km-wide viewing swath, MODIS however sees every point on our world every 1–2 days in 36 discrete spectral bands. Consequently, MODIS tracks a wider array of the earth's vital signs than any other Terra sensor. For instance, the sensor measures the percent of the planet's surface that is covered by clouds almost every day. This wide spatial coverage enables MODIS, together with MISR and CERES, to help scientists determine the impact of clouds and aerosols on the Earth's energy budget. MODIS provides daily global coverage and the 10 km resolution of aerosol optical depth (AOD) for studying spatial variability of aerosols in urban areas. AOD is a reasonably good proxy for PM2.5 ground concentrations. Wang et al. [19] used this data to generate air quality maps and compared the information to the acute health needs of the residence in the test area. Téllez-Rojo et al. [16] used the same air-quality mapping approach for Mexico City and showed the direct relation of low air qualities to the acute respiratory symptoms of children.

The MOPITT sensor was designed to enhance our knowledge of the lower atmosphere and to observe how it interacts with the land and ocean biospheres. The sensor measures emitted and reflected radiance from the Earth in three spectral bands. As this light enters the sensor, it passes along two different paths through onboard containers of carbon monoxide. The different paths absorb different amounts of energy, leading to small differences in the resulting signals that correlate with the presence of these gases in the atmosphere. MOPITT's spatial resolution is 22 km at nadir and it "sees" the Earth in swaths that are 640 km wide. Moreover, it can measure the concentrations of carbon monoxide in 5 km layers down a vertical column of atmosphere, to help scientists track the gas back to its sources.

Another highly used data source comes from the Landsat satellite (passive) sensors. Launched in 1982 on Landsat 4 and 1984 on Landsat 5, researchers found an opportunity to access the TM thermal bands. Landsat 4 operated successfully for over 10 years, with data collection terminated in 1993. The Landsat 5 TM acquired data for over 27 years until communication system failures essentially ended the TM data collections in November 2011. Landsat 6 never reached its operational orbit after launching in 1993. In 1999, Landsat 7 was launched with the ETM+ instrument. The newest Landsat mission, Landsat Data Continuity Mission (LDCM, or Landsat 8 after launch), was launched in February 2013 carrying the next-generation Landsat thermal-imaging sensor.

The Landsat TM data is one of the most widely used satellite images for LST retrieving because of its high resolution (120 m) and free download availability from the website of US Geological Survey (USGS), which has one thermal infrared (TIR) band. This makes retrieving LST from a single band more difficult than from multiple thermal bands. In comparison, ASTER data has five thermal bands with a higher resolution (90 m), which may provide more promising potential for LST retrieval studies, although very few studies of LST retrieval from ASTER data are available as yet. Therefore, in this study, we applied the mono-window algorithm to the Landsat TM and the split-window algorithm to ASTER data for the analysis of its effect of urban heat island in the case study of Hong Kong. Although satellite data (e.g., Landsat TM and ASTER thermal bands data) can be applied to examine the distribution of urban heat islands in places such as Hong Kong, the method still needs to be refined with in situ measurements of LST in future studies. Among others, Ho et al. [9] used Landsat ETM+ to map urban temperature and compared the results to the local weather station results. This comparison shows the high accuracy of the satellite based observations.

Table 1 compares the spatial resolutions of the most frequently used passive sensors. Higher resolution thermal images could also be obtained from airborne images. Even though the airborne images provide very high-resolution information, it requires a very expensive procedure to fly over a large area to collect information. If the observation needs to be repeated with a certain time frequency, this procedure opens technical and financial challenges. Therefore, satellite sensors are highly preferred both by researchers and other land-observation institutions.

### 3.3 Further Supportive Data that Satellite Remote Sensing Can Offer

Besides providing measurements about wind speed/direction, air quality, and the heat islands, satellite sensors can make more measurements about earth which could support detailed research in the field of urban airflow analysis. For instance, multispectral satellite images provide an opportunity to calculate normalized difference vegetation index (NDVI) which tells about the vegetation greenness in the area. Some researchers have used NDVI as a major indicator of the urban climate. Experiments showed that during summer-time NDVI is negatively correlated with surface temperature. However, previous studies have also shown a causal relationship between NDVI and LST that is subject to seasonal variation. Furthermore, the response of LST to NDVI varies among different land cover types. Some studies have demonstrated that LST has a stronger correlation with other parameters which could be obtained by using multispectral satellite image bands. For instance, the normalized difference built-up index (NDBI) or vegetation fraction than with NDVI showed higher correlation to the urban LST.

**Table 1** Remote-Sensing Satellite Sensors for measuring LST and their spatial resolutions

| Remote-sensing sensor | Spatial resolution (meter) |
|---|---|
| ASTER | 90 |
| Landsat 3 MSS | 240 |
| Landsat 4, 5 TM | 120 |
| Landsat 7 ETM+ | 60 |

Impervious surfaces are found in urban and suburban landscapes and can be related to population density and urbanization. It is shown that reliable quantification of UHI could be achieved by analyzing the relationship between LST and impervious surface areas, which are again calculated by using multispectral bands of the satellite images.

Besides providing vegetation and built-up area indicator parameters from the multispectral band ratios, satellite images also offer the opportunity of automatic mapping of detailed 2D or 3D structures and for observing the traffic density as well. In a previous study, Sirmacek et al. [12] have proposed a 3D building reconstruction method compared to the rooftop shape prediction accuracies when different satellite sensors are used. Huang et al. [10] have studied impacts of different building roof shapes on the urban airflow. Their study shows that automatic building detection and rooftop modeling methods might provide important information to enhance the urban air quality and airflow research. Taubenboeck et al. [15] have studied extracting patterns from satellite images which can indicate social groups and income distribution within urban areas. These studies might also provide more discussion points to the urban air quality analysis research. Last but not least, there is probably a significant correlation between road usage and air quality in urban areas. Zheng et al. [23] used AI to estimate urban air quality. They trained their models with features coming from satellite data, which indicate distances to the road network, length of the roads, and meteorological data. Their study showed the significant correlation of the urban air quality with the extracted road features. Therefore, both the passive sensor based information and active sensor based information might be important satellite sensor based sources to study the urban air quality.

## 4    Challenges and Open Problems

In the previous sections, we have seen that the application of machine-learning methods can be helpful in the study of urban air pollution. Two approaches were covered, that is, air-pollution models – which consisted on directly modeling air pollution – and flow-dynamics models, which focused on inferring the behavior of pollutants from the dynamics of the flow. Nevertheless, both approaches exhibit limitations and areas of improvement. On the one hand, the application of machine-learning algorithms to air-pollution models requires large amounts of data to provide accurate predictions. On the other hand, modeling the dynamics of the flow, despite not needing such databases, requires running numerical simulations which have an important computational cost. Furthermore, the inference of pollutant dispersion from the dynamics of the flow is not direct and thus requires a proper validation.

Additionally, another challenge affecting both approaches is the training of networks. It is known that the network training and the employed data usually have an important influence on the outcome of the system. In fact, the matter of training biases and, by extension, the biases of the system are important questions in all the fields where artificial intelligence is applied. The assessment of this problematic is relevant both from a technical perspective – since it heavily affects the outcome of the predictions produced by the network – and from an ethical point of view, since the outcome of those networks will typically influence policy-making decisions.

On the other hand, active sensors can help with direct measurements of the wind speed and direction. However, due to their low resolutions, their application areas are limited with sea and ocean monitoring. Their resolutions cannot help with adding valuable information to study urban airflow. On the other hand, passive sensors of the satellites provide measurements which can indirectly help with the urban airflow studies. They can measure air quality, local heat, vegetation density and greenness. They can even provide detailed 3D building models and information about the road network. Even though these parameters are highly correlated with urban airflow, for precise mapping, still it is important to calibrate and enhance the satellite based information with in situ sensors. Budde et al. [2] introduced SmartAQnet called study which aims to combine multiple data resources to study the urban air quality in detail. They have combined IoT data such as weather sensors, in-situ air quality sensors, dust measurement devices, satellite images, and high-resolution drone images. However, due to the scalability issue of such a complex sensor fusion challenge, they have limited their analysis only with the Augsburg city of Germany. This study shows the big data collection and analysis challenges which still exist.

Another challenge is the validation of the health impacts of the urban airflow analysis. Even though the earlier studies showed a correlation with respiratory diseases, further research is still needed to completely pinpoint the urban-airflow aspects responsible for it. This will allow to develop better residence health estimation models, for increased urban sustainability.